\documentclass[10pt,twocolumn,letterpaper]{article}

\usepackage{iccv}
\usepackage{times}
\usepackage{epsfig}
\usepackage{graphicx}
\usepackage{amsmath}
\usepackage{amssymb}
\usepackage{amssymb}
\usepackage{subcaption}
\usepackage{float}

\usepackage[breaklinks=true,bookmarks=false]{hyperref}

\iccvfinalcopy 

\def\iccvPaperID{9} 

\ificcvfinal\fi
\begin{document}
\pagenumbering{gobble}

\title{Exploiting Convolution Filter Patterns for Transfer Learning}

\author{Mehmet Ayg\"{u}n\\
Istanbul Technical University\\
Istanbul, Turkey\\
{\tt\small aygunme@itu.edu.tr}
\and
Yusuf Aytar\\
MIT\\
Cambridge, USA\\
{\tt\small yusuf@csail.mit.edu }
\and
Haz{\i}m Kemal Ekenel\\
Istanbul Technical University\\
Istanbul, Turkey \\
{\tt\small ekenel@itu.edu.tr}
}

\maketitle

\begin{abstract}
In this paper, we introduce a new regularization technique for transfer learning. The aim of the proposed approach is to capture statistical relationships among convolution filters learned from a well-trained network and transfer this knowledge to another network. Since convolution filters of the prevalent deep Convolutional Neural Network (CNN) models share a number of similar patterns, in order to speed up the learning procedure, we capture such correlations by Gaussian Mixture Models (GMMs) and transfer them using a regularization term. We have conducted extensive experiments on the CIFAR10, Places2, and CMPlaces datasets to assess generalizability, task transferability, and cross-model transferability of the proposed approach, respectively. The experimental results show that the feature representations have efficiently been learned and transferred through the proposed statistical regularization scheme. Moreover, our method is an architecture independent approach, which is applicable for a variety of CNN architectures.
\end{abstract}

\section{Introduction}
\label{sec:intro}
The CNN models are found to be successful at various computer vision tasks such as image classification \cite{he2016deep,simonyan2014very,szegedy2015going}, object detection \cite{redmon2016you}, image segmentation \cite{long2015fully,pinheiro2015learning}, and face recognition \cite{parkhi2015deep}, where large-scale datasets \cite{russakovsky2015imagenet,zhou2016places,parkhi2015deep} are available. Nevertheless, the performance of CNN models significantly reduces, when training data is limited or the domain of the training set is far from the test set. Today, the most successful and practical solution to address lack of annotated large dataset is training the networks on large-scale annotated datasets like ImageNet \cite{russakovsky2015imagenet} and Places \cite{zhou2016places}, then finetuning these pre-trained networks for specific problems. Thanks to community, the pre-trained models of well-known architectures like AlexNet \cite{krizhevsky2012imagenet}, VGG-16 \cite{simonyan2014very}, GoogLeNet \cite{szegedy2015going}, and ResNet \cite{he2016deep} can be found available online. However, when some architectural changes are needed, these pre-trained networks cannot be used. For such cases, it is necessary to train models on large datasets, then, finetune for the particular problem. Unfortunately, while getting nearly human performances on a lot of applications with these models, training these networks on large datasets is still a significant problem and a very time consuming process.
\par
\begin{figure*}
\begin{center}
\includegraphics[width=0.8\linewidth]{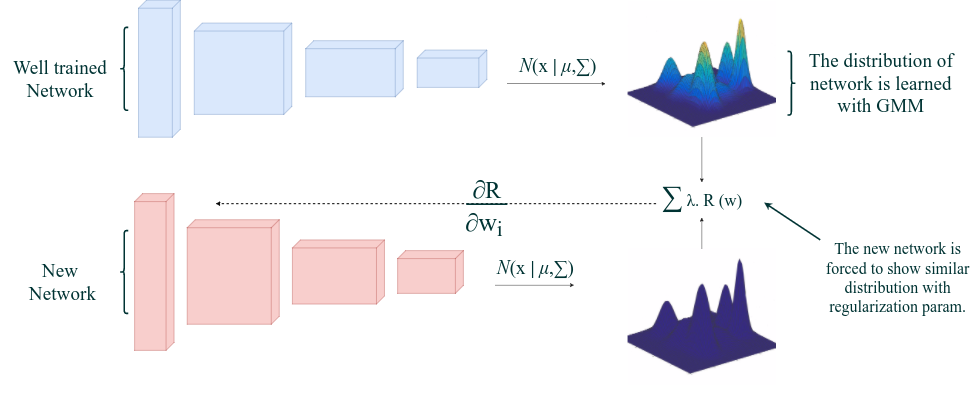}
\end{center}
   \caption{Overview of the proposed system. The blue (top) network is a well-trained CNN where its distribution is learned via GMM. When  new red (bottom) network  being trained, along with the classification loss, the new regularization term $\sum \lambda \cdot R(w)$  that measures the statistical difference of two distributions are minimized. The statistical knowledge could be transferred from source CNN to target CNN with our regularization term. Best viewed in color. }
\label{fig:main_method}
\end{figure*}

 With recent advances in deep learning such as Residual Learning \cite{he2016deep}, successful networks have become more and more deep, and training these models become harder in terms of complexity and time. To find a solution for this problem, inspired by \textit{What makes a good detector?} \cite{gao2012makes}, we have investigated \textit{What makes a good CNN filter?} Two successful CNN models, AlexNet \cite{krizhevsky2012imagenet} and VGG-16 \cite{simonyan2014very}, are analyzed from a statistical perspective, and we realized that these models show similar patterns and redundancies. In addition to our findings, in \cite{denil2013predicting}, authors show that 95\% of weights of neural networks could be predicted without any reduction in accuracy. This leads us to the idea that we can use these redundancies and patterns for learning better representations quickly by transfer learning. Similar to the methods used in \cite{Aytar15,Castrejon16,gao2012makes}, we introduce a regularization term for transferring the statistical information to improve the learning scheme. First, the statistical distribution of convolution filters from a well-trained network is learned with a Gaussian Mixture Model. Next, the newly trained model is encouraged to show similar statistics with source models using the regularization term. Extensive experiments on the CIFAR10 \cite{krizhevsky2009learning}, Places2 \cite{zhou2016places}, and CMPlaces \cite{Castrejon16} datasets show that the proposed approach is generalizable and the networks can quickly learn a representation with statistical regularization, which could efficiently be transferred to another task and cross-domains. The overview of our proposed method can be seen in Figure \ref{fig:main_method}. The rest of the paper is organized as follows: in Section \ref{rwork} related works are summarized, in Section \ref{method} our detailed statistical analysis is reviewed and the regularization term is introduced, in Section \ref{experiments} experimental results are presented and discussed, and finally in Section \ref{conclusion} the paper is concluded.

\section{Related Work}
\label{rwork}

\textbf{Network Distillation: }The aim of network distillation approaches are transforming larger networks to smaller ones, while not losing so much information that was learned in the large network. The pioneering work is conducted by Bucila \etal \cite{buciluǎ2006model}, where their aim is to compress the ensemble of models to a single model without significant accuracy loss. Later, Hinton \etal \cite{hinton2015distilling} optimized the smaller network to show similar softmax output of cumbersome model. Then, Romero \etal \cite{romero2014fitnets} suggest that in addition to softmax outputs, intermediate representations could be used for distilling the network. Recently distillation is also applied in the cross-modal settings \cite{gupta2016cross,aytar2016soundnet}. The major drawback of these methods is the necessity to train a large network before using it to train a smaller one. Also these models mainly match the outputs of the networks, whereas we regularize the internal weights of the network. \par 
\textbf{Domain Adaptation: } Domain adaptation is the problem of learning a model that generalizes target domain examples besides source domain ones while learning only from source domain. With the success of CNNs, domain adaptation works have been focused on CNNs and several successful methods have been proposed. For instance, Ganin \& Lempitsky \cite{ganin2014unsupervised} introduced \textit{gradient reversal layer} which act like identity transform in forward pass of CNN and change the sign of the gradient and scale in the backward pass. In their work, they added a domain classifier to the end of the feature map and tried to predict the domain of examples. In backpropagation, they changed the gradient using gradient reversal layer and forced the networks to learn domain invariant features by maximizing loss of domain classifier. Tzeng \etal \cite{tzeng2015simultaneous} added new terms into objective function of CNN to both increasing domain confusion and transferring inter-class knowledge. The first term \textit{domain confusion loss} forces the networks to learn domain invariant features and \textit{soft label loss} forces the feature of the same class to be similar for both source and target domain. In contradistinction to Ganin \& Lempitsky \cite{ganin2014unsupervised}, some target labels must be available for optimizing soft label loss. Moreover, recently some other works \cite{ganin2016domain,ghifary2016deep,tzeng2017adversarial,sener2016unsupervised} focused on domain adaptation problems for CNNs. While these methods focus on transferring information about structure of data, our method focuses on transferring more local information.\par \textbf{Statistical Transfer: } Statistical transfer is learning statistical properties from a source and to use this statistical knowledge for improving learning procedure. For instance, Aytar \& Zisserman \cite{Aytar15} proposed part-level transfer regularization which transfers parts of source detectors instead of whole detector. Additionally, they take advantage of part co-occurrence statistics. For example, if there are wheels in the picture, probably another wheel would also appear. They calculated these co-occurrence statistics using the source data and transfer these statistics when a new object detector is learned. Moreover, in the era of hand crafted features, Gao \etal \cite{gao2012makes} analyzed famous HOG \cite{dalal2005histograms} templates of successful object detectors and made two observations. Firstly, they observed that activations of individual cell models had some correlations, secondly local neighborhoods of cells also showed the same characteristic. Furthermore, since they wanted to transfer local information in contrast to templates like in \cite{Aytar15}, they defined their priors such that the correlations from the source model could be transferred to target model without global template alignment. In a recent work \cite{Castrejon16}, GMMs are used for aligning cross-modal data. In this approach, statistics of activation maps of different layers for a modality are learned, and the other modalities are forced to show similar statistics in the activations via a regularization term. The proposed method is capable of aligning modalities by using this regularization term when there is no strong alignment between modalities.\par Our work is influenced by Gao \etal \cite{gao2012makes} and Castrejon \etal \cite{Castrejon16}. The first work directs us to analyze the weights of convolution filters and find the correlation between the filters, and the other one to use the GMM to enforce them to show similar statistics in a non-convex optimization problem. Different from these works, our regularization forces the weights to show similar statistics and transfers local correlations of convolution filters. 


\section{Statistical Regularization}
\label{method}
In  this  section, first, we present our analysis on CNN models from a statistical perspective. Next,  we describe our approach for capturing statistical knowledge from a CNN and present how to transfer this knowledge via a regularization term.

\begin{figure*}[t!]
    \centering
    \begin{subfigure}[t]{0.54\linewidth}
        \centering
        \includegraphics[width=0.85\linewidth]{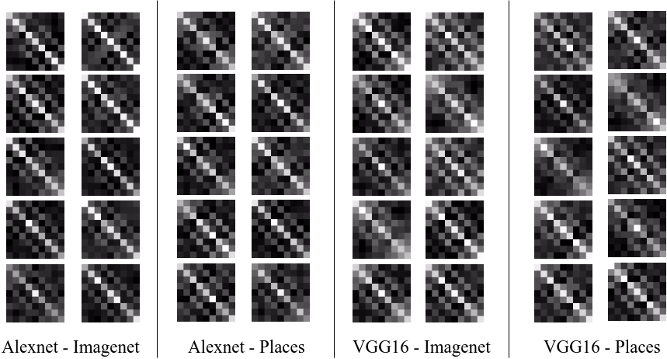}
        \caption{}
    \end{subfigure}%
    ~
    \begin{subfigure}[t]{0.41\linewidth}
        \centering
        \includegraphics[width=0.85\linewidth]{means.png}
        \caption{}
    \end{subfigure}
    \caption{(a) The covariance matrices of ten clusters that are calculated using the weights of convolution filters. The converged VGG-16 and AlexNet models, which are trained on Places and ImageNet datasets, are used for clustering and visualizing. (b) Visualizations of mean values of each clusters.}
\label{fig:cov-mean}
\end{figure*}

\subsection{Statistical Analysis}
\begin{figure*}
\begin{center}
\includegraphics[width=0.80\linewidth]{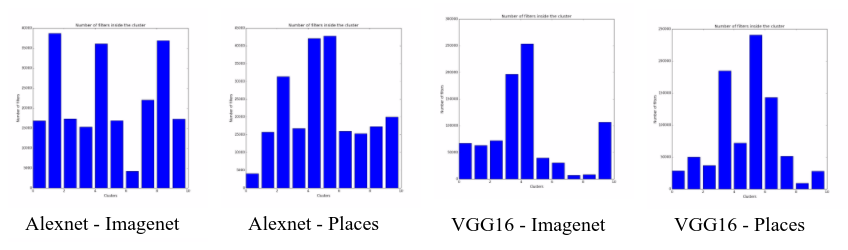}
\end{center}
   \caption{The distributions of the convolution filters over clusters. For both Places and ImageNet, the converged model of VGG and AlexNet show similar distributions.}   
\label{fig:distibutions}
\end{figure*}

For investigating the statistical behavior of a CNN model, we have used VGG-16 \cite{simonyan2014very} and AlexNet \cite{krizhevsky2012imagenet} models that were trained on ImageNet \cite{russakovsky2015imagenet} and Places-365 \cite{zhou2016places} datasets. Especially, weights of convolution filters are analyzed for the four models. We tried to answer following three questions:  (i)  are filters separable into clusters?; (ii) how much similar are the filters inside a cluster?; and (iii) how all filters are distributed over clusters?. In order to obtain this information, all $3 \times 3$ filters of a model are clustered to ten different clusters using the k-means algorithm and each cluster's covariance matrix and mean value were calculated and visualized. The covariance matrices would provide information about how members of a cluster are correlated. Mean values show whether the sets are similar to each other or not. For instance, it can be seen in Figure \ref{fig:cov-mean} (a) the covariance matrices of all the four models show that there is some shared behavior of learned filters across layers and architectures. Also, the cluster centers depict different similarities. Generally, all models have clusters that their mean values are accumulated on the left, right, top, and bottom. Moreover, when we look at the distribution of filters to clusters, the models show different characteristics. While the distributions are Gaussian-like in VGG-16 models, the distributions cannot be fitted to a known distribution in the AlexNet models and most of the clusters have roughly the same number of filters. However, both AlexNet and VGG-16 models show similar distributions by model wise while they are trained on different dataset. The mean values are shown in Figure \ref{fig:cov-mean} (b), and the distributions can be seen in Figure \ref{fig:distibutions}.


\subsection{The Proposed Method}
Since our aim is capturing statistical properties of ``good" convolution filters and transfering this statistical knowledge to another network, we modeled the convolution filters' distributions and enforced the network to show similar distributions. Similar to our work, \cite{Castrejon16} forces the activations of networks to show similar distributions across modalities for aligning cross modal data. In \cite{Castrejon16}, the authors used both mixture and single Gaussian distributions for modelling activations, and the mixture models outperformed single Gaussian model in their problem. Since, our aim is also similar to their work -capturing statistical knowledge and transfering it- we have decided to use mixture models for modeling distributions. The main difference is that we force the weights of filters instead of the activations to show similar distributions across the networks.\\
Let $x_n$ and $y_n$ be a training image and its corresponding label. We want to minimize  
\begin{equation}
\underset{w} {\mathrm{min}}  \sum_{n} L (z(x_{n},w),y_{n})
\end{equation}
where $z_n$($x_n$ ,w) is output of the network. We have added a regularization term $R$ to the loss term that represents the negative log
likelihood of a convolution filter to encourage the network to learn the weights that are statistically similar to another network. For filter $w_i$ and distribution $P$ we define $R$ such that,
\begin{equation}
R(w_{i}) = -\log(P(w_{i})) 
\end{equation}
The distribution $P$ is modeled as GMM, therefore  $P$  would be,
\begin{equation}
P(w | \pi, \mu, \Sigma) =  \sum_{k=1}^{K} \pi_{k} \mathcal{N} (w | \mu_{k},\Sigma_{k})
\end{equation}
where $K$ is the number of mixtures and $\sum_{k} \pi_{k}=1$ , $\pi_{k}  \geq 0 $ $ \forall_{k} $. Therefore, the total negative log likelihood for $N$ convolution filters can be defined as,
\begin{equation}
R(w | \mu, \Sigma) = \sum_{i=1}^{N} -\log \sum_{k=1}^{K} \pi_{k}  \mathcal{N} (w_{i} | \mu_{k},\Sigma_{k})
\end{equation}
where $\mathcal{N}$ is specified as,
\begin{equation}
\mathcal{N}( w | \mu, \Sigma)= \frac{1}{(2\pi|\Sigma|)^-(\frac{1}{2})}\exp( -\frac{1}{2} (w-\mu)^{T}\Sigma^{-1} (w-\mu))
\end{equation}
The derivate of the negative log likelihood must be calculated exactly or analytically, since the network is trained with back-propagation. Still, calculating exact derivative in every iteration would be very expensive during training, therefore, we approximately calculate the derivative. To approximately calculate the derivative of a convolution filter, we first pick the mixture component $\mathcal{N} (\mu_{s},\Sigma_{s})$ that the probability of the convolution filter is maximized. Next, the derivate is calculated using that single Gaussian. The partial derivative of $R$ with respect to a convolution filter $w_{i}$ would be  
\begin{equation}
\frac{\partial R}{\partial w_{i}}
   = (w_{i}-\mu_{s})\Sigma_{s}^{-1} 
\end{equation}
Finally, our complete loss term is defined as
\begin{equation}
\min_{\textrm{w}}\limits \sum_{n} L (z(x_{n},w),y_{n}) + \sum\ \alpha \frac{1}{2} w^2 + \sum \lambda \cdot  R(w)
\end{equation}
where the first and second terms are classification loss and weight decay terms, and the last one is our regularization term. The $\lambda$ is a hyperparameter that controls the regularization. For experiments in this paper, we have used distribution $P$, which is learned from convolution filters of VGG-16 model trained on ImageNet using Expectation maximization (EM) algorithm with $K=1000$ components. To reduce the number of parameters, we assumed that the covariances $\Sigma_{k}$ are diagonal. Also, K-Means algorithm was run on filters for decreasing the convergence time of EM algorithm. While it is known that, different layers, especially first layers, show different characteristics than other layers, we have used a single GMM -$P$- for all filters. Because, if multiple distributions are used, a layer alignment relationship between layers are needed in transfer time.
\begin{figure*}[t!]
    \centering
    \begin{subfigure}[t]{0.23\linewidth}
        \centering
        \includegraphics[width=0.999\linewidth]{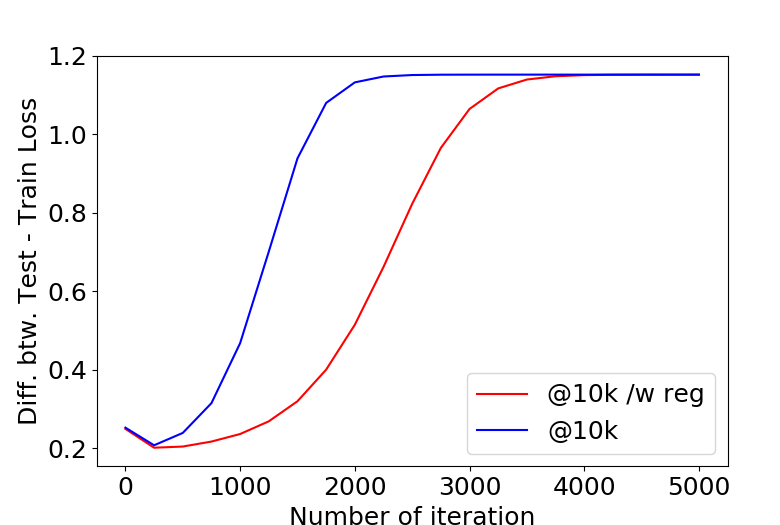}
        \caption{}
    \end{subfigure}%
    ~
    \begin{subfigure}[t]{0.23\linewidth}
        \centering
        \includegraphics[width=0.999\linewidth]{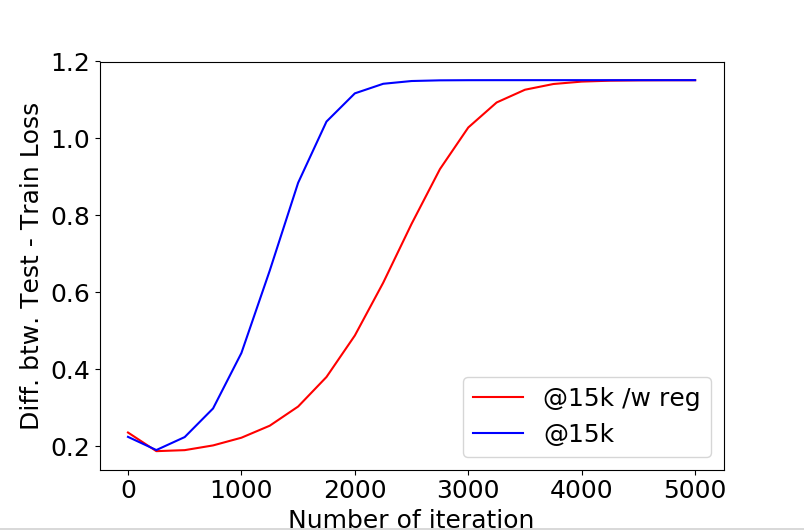}
        \caption{}
    \end{subfigure}%
    ~
    \begin{subfigure}[t]{0.23\linewidth}
        \centering
        \includegraphics[width=0.999\linewidth]{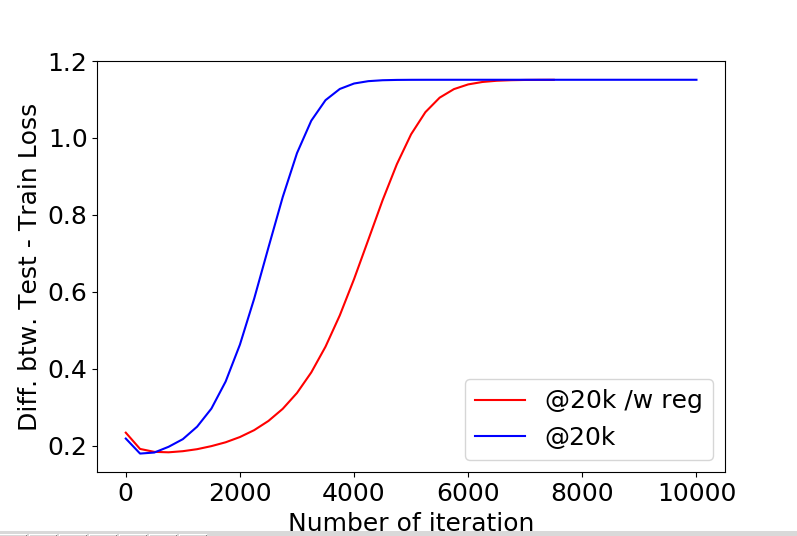}
        \caption{}
    \end{subfigure}%
    ~
     \begin{subfigure}[t]{0.23\linewidth}
        \centering
        \includegraphics[width=0.999\linewidth]{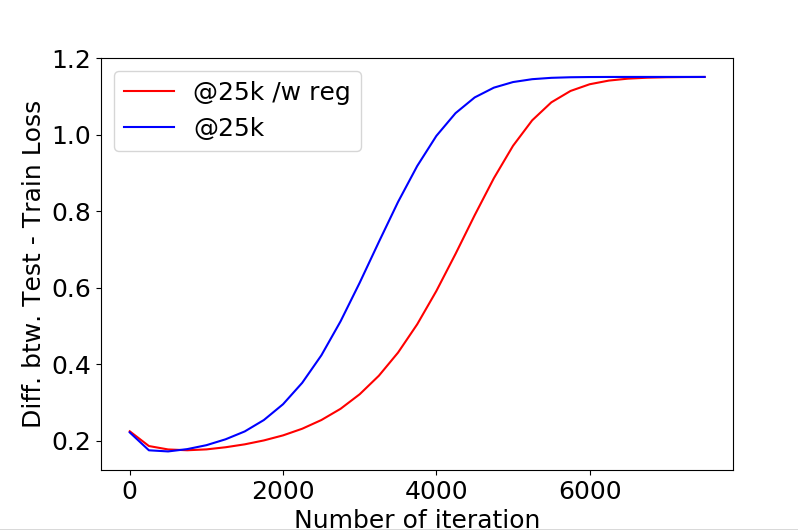}
        \caption{}
    \end{subfigure}%
    
   \caption{Difference between train and test losses of models in CIFAR-10. After freezing weights the models continued to training. It can be easily seen that later freezing show less overfitting. In the (a),(b),(c), and (d) the regularization effect is compared. With regularization models always overfit slowly (the difference increases slowly) than regular training. Best viewed in color.}
\label{fig:cifar}
\end{figure*}
\section{Experiments and Results}\label{experiments}
Our hypothesis was that well-trained CNN models show similar statistical patterns, and we could exploit this information in the training phase. Our experiments show that we can learn a better representation faster with the statistical regularization. For example, in section \ref{exp1}, we validate that with regularization the convolution filters create more general representations and show less overfitting characteristics than regular trained filter's representations. Also in section \ref{exp2}, we see how the regularization helps to learn representations that are successfully transferred to another task. Furthermore, while transferring across task and domains is widely studied, transfer learning for cross modal data is not a well studied problem. We show that our method could be applied to cross-modal data as well. In section \ref{exp3}, we present our experiments applied to a cross-modal dataset CMPlaces \cite{Castrejon16}.

\subsection{Generalization} \label{exp1}

In this section, we evaluate if the statistical regularization helps the generalizability of the learned representation. For this purpose, we have used middle sized CIFAR-10 dataset \cite{krizhevsky2009learning} and CNN architecture described in \cite{salimans2016weight}. Firstly, we trained the network with and without regularization and stopped the training at 10k, 15k, 20k, and 25k iterations. Later on we freeze all $3 \times 3$ filters and continue  learning with training data. Since only the last layers change during the training and the features extracted from convolution filters are not so generic, the validation loss starts to increase and the validation accuracy is dropped.

\par We compared the networks, whose training stopped in different iterations with and without regularization. In our experiments, the training loss is started to oscillate in a small interval and does not change much, that is, neither increases nor decreases. However, the validation loss changed. The gap between training and test loss indicates how generalizable the network is.

When we compared the models initialized at 10k, 15k, 20k, and 25k iterations, always the test loss of regularized versions increased slowly compared to those of non-regularized versions. For instance, as can be seen in Figure \ref{fig:cifar} (a,b,c,d), the difference between training and test loss increases slowly in regularized models than non-regularized models. Interestingly, as regular training time (no freezing) increases, the gap between the regularized and normal training reduces.

\subsection{Task Transfer}\label{exp2}
\begin{figure*}[t!]
    \centering
    
    \begin{subfigure}[t]{0.32\linewidth}
        \centering
        \includegraphics[width=0.75\linewidth]{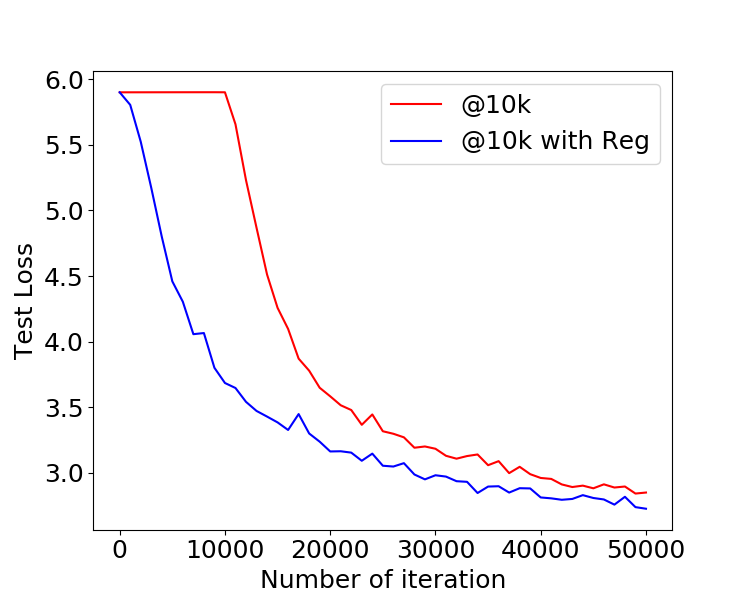}
        \caption{}
    \end{subfigure}%
    ~
    \begin{subfigure}[t]{0.32\linewidth}
        \centering
        \includegraphics[width=0.75\linewidth]{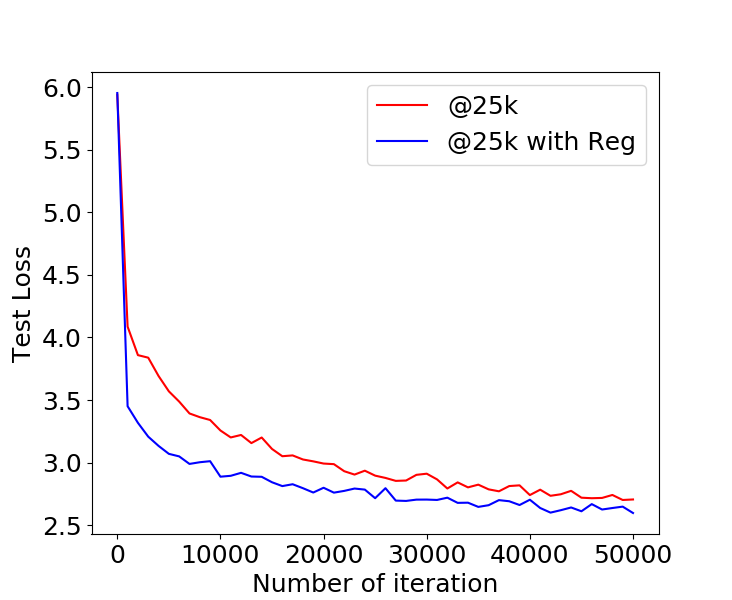}
        \caption{}
    \end{subfigure}
    ~
    \begin{subfigure}[t]{0.31\linewidth}
        \centering
        \includegraphics[width=0.75\linewidth]{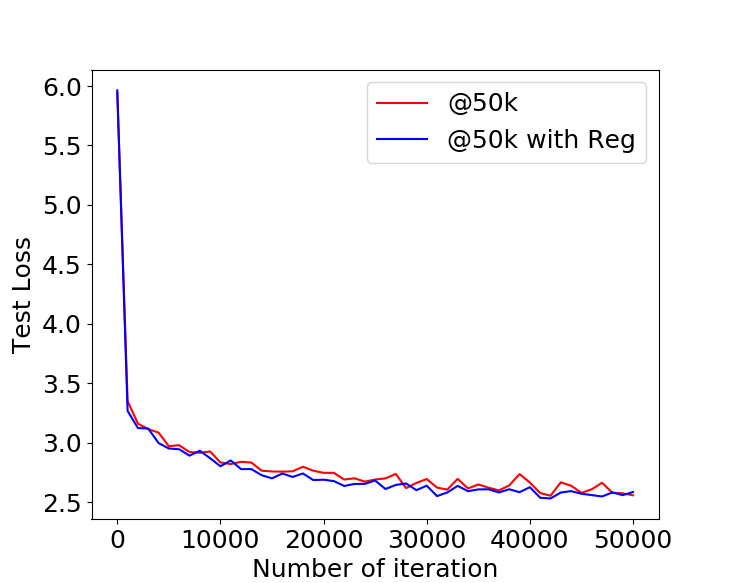}
        \caption{}
    \end{subfigure}%
   \caption{The pre-traiened VGG-F models are finetuned on Places2 dataset. Pretraining is stopped at 10k, 25k, and 50k iterations. For (a) and (b) it can be easily seen that with regularization the test loss decreases quickly. Since test losses and accuracies correlated we only provide test losses.}
\label{fig:places}
\end{figure*}

In this section, we show that the quickly learned representations can also be transferred to another task more successfully with regularization. To validate our claim, we try to transfer filter distributions from ImageNet to Places2 \cite{zhou2016places} dataset. The trivial solution for a classification problem is training the network on ImageNet and finetuning the models on the new task afterwards. We follow the same procedure in this section for our experiments, but we want to finish the pre-training stage as early as possible. Also we want to show that our regularization can be used with finetuning. To evaluate our method's performance, we first train VGG-F model introduced in \cite{Chatfield14} on ImageNet \cite{russakovsky2015imagenet} data with and without regularization. As in Section \ref{exp1}, we take snapshots from 10k, 25k, and 50k iterations. Next, we start finetuning on Places2 data \cite{zhou2016places}. When we compare regularization effect, we see that the regularization can help to learn better representations in the early iterations. For example, as can be seen in Figure \ref{fig:places} (a), the performance of the models that are initialized at the weights learned in pre-training with only 10k iterations, the regularized version shows a better performance than the non-regularized one. When we examine the models initialized at 25k iterations, the performance difference between regularized and non-regularized versions reduces. Finally, in 50k iterations, nearly there is no difference in the performance. The test loss/iterations plots are shown in Figure \ref{fig:places} (a), (b), (c). This experiment shows that as pre-training time increases, the gain obtained from regularization decreases. However, for limited amount of pre-training time, the regularization could increase the efficiency of the pre-training. 

\subsection{Modality Transfer}\label{exp3}
While CNNs performances are very good at various computer vision tasks for real-world images, most of the computer vision algorithms fail at non-real images. This shows that generalization performance of the computer vision algorithms is not good for cross modal data. Some recent works \cite{Castrejon16,eitz2011sketch} focused on this problem and recently Castrejon \etal \cite{Castrejon16} has introduced a new Cross-Modal dataset. In the dataset there are five different modalities for each scene type such as natural image, sketch, clip art, spatial text, and description. As in section \ref{exp2}, we firstly train VGG-F models on a large dataset -Places2- and finetune on CMPlaces data. We have used clipart and sketch data to evaluate our performance, since we are interested in the visual domain. We take snapshots from 25k and 50k iterations from the regularized and non-regularized networks. Next, we start to finetune on sketch and clipart data and compare their accuracies. Also, we finetune VGG-F model converged on ImageNet data and compare with our pre-trained models. Similar to our experiments, by rising the training pre-training time the performance increases. Also, the regularization helps to learn better representations and increases the performance of pre-training for finetuning. Although, there is a significant gap between the converged ImageNet model and our regularization in the sketch data, there is not a substantial difference between 50k iteration with regularization and the converged ImageNet model on clipart modality. The results for both modalities can be seen in Table \ref{table:table1} and Table \ref{table:table2}. These results show that instead of training ImageNet until the model converges, we could train the models using regularization with only few iterations and could employ these pre-trained networks for cross-modality transfer. 
\begin{table}
\begin{center}
\begin{tabular}{|l|c|}
\hline
Pre-Training & Top-5 Accuracy \\
\hline\hline
25k & 60.45 \\
25k \textbackslash w Reg. & 62.25 \\
50k & 63.0 \\
50k \textbackslash w Reg. & 64.5\\
Converged Imagenet & 64.8\\
\hline
\end{tabular}
\caption{The first column describes how pretraining is done and the second column shows the top-5 accuracies after finetuning for clipart data.}
\label{table:table1}
\end{center}
\end{table}

\begin{table}
\begin{center}
\begin{tabular}{|l|c|}
\hline
Pre-Training & Top-5 Accuracy \\
\hline\hline
25k & 33.05 \\
25k \textbackslash w Reg. & 40.75 \\
50k & 37.65 \\
50k \textbackslash w Reg. & 41.1\\
Converged ImageNet & 53.6\\
\hline
\end{tabular}
\caption{The first column describes how pretraining is done and the second column shows the top-5 accuracies after finetuning for sketch data.}
\label{table:table2}
\end{center}
\end{table}

\subsection{Implementation Details}\label{exp4}
We have used Caffe \cite{jia2014caffe} deep learning framework in our experiments. Moreover, we have implemented our special convolution layer for applying statistical regularization. When VGG-F model is trained on ImageNet and Places datasets, stochastic gradient descent with 0.01 learning rate is used for optimization. In the CIFAR experiments, we have used the same parameters described in \cite{salimans2016weight}. Finally, the Gaussian mixture models are learned using the VLFeat library \cite{vedaldi2010vlfeat}.
\section{Conclusion}
\label{conclusion}
In this paper, we analyzed convolution filters of well-known CNN architectures and found that they share a number of common patterns and redundancies that could be exploited for transfer learning. Gaussian Mixture Models are used for capturing these statistical patterns and a new regularization term is introduced for transferring such patterns to other networks. Our experiments show that we could learn good representations that are transferable to the other tasks and cross-domains quickly with regularization. For instance, we achieved around 25\% improvement on the sketch modality in the cross-modal dataset under limited pre-training time. Also, our method gets similar performance on clipart data with converged model that pre-trained on the ImageNet, while pre-training stopped at 50k iterations in our method.

{\small
\bibliographystyle{ieee}
\bibliography{egbib}
}

\end{document}